# ACCELERATED AIRFOIL DESIGN USING NEURAL NETWORK APPROACHES

**Anantram Patel**[1, †]**, Nikhil Mogre**[2, †]**, Mandar Mane**[3]**, Jayavardhan Reddy Enumula**[2]**,
Vijay Kumar Sutrakar**[2, *]

[1]SCOPE College of Engineering, Bhopal, India
[2]ADE, DRDO, Bengaluru, India
[3]Indian Institute of Technology Hyderabad, TG, India

**ABSTRACT**

*In this paper, prediction of airfoil shape from targeted pressure distribution (suction and pressure sides) and vice versa is demonstrated using both Convolutional Neural Networks (CNNs) and Deep Neural Networks (DNNs) techniques. The dataset is generated for 1600 airfoil shapes, with simulations carried out at Reynolds numbers (Re) ranging from 10,000 and 90,00,000 and angles of attack (AoA) ranging from 0 to 15 degrees, ensuring the dataset captured diverse aerodynamic conditions. Five different CNN and DNN models are developed depending on the input/output parameters. Results demonstrate that the refined models exhibit improved efficiency, with the DNN model achieving a multi-fold reduction in training time compared to the CNN model for complex datasets consisting of varying airfoil, Re, and AoA. The predicted airfoil shapes/pressure distribution closely match the targeted values, validating the effectiveness of deep learning frameworks. However, the performance of CNN models is found to be better compared to DNN models. Lastly, a flying wing aircraft model of wingspan >10 m is considered for the prediction of pressure distribution along the chordwise. The proposed CNN and DNN models show promising results. This research underscores the potential of deep learning models accelerating aerodynamic optimization and advancing the design of high-performance airfoils.*

**Keywords: Airfoil, Convolutional Neural Networks (CNN), Deep Neural Networks (DNN), Reynold number ($Re$), Angle of attack ($AoA$)**

## 1. INTRODUCTION

Airfoil design is fundamental to the performance of aerodynamic systems, particularly in the aerospace and aviation sectors. Optimizing airfoil shapes to meet performance requirements has traditionally relied on Computational Fluid Dynamics (CFD) simulations and experimental testing. Although highly accurate, these methods are resource-intensive, requiring significant computational power and time for iterative optimization processes. To address these limitations, the rise of Artificial Intelligence (AI), particularly deep learning, has brought about promising alternatives that can reduce computational cost and improve de- sign efficiency [1-11]. CNN and DNN have shown great potential in predicting aerodynamic performance from geometric data. These approaches leverage large datasets to learn complex relationships between airfoil shapes and their aerodynamic characteristics, such as lift and drag coefficients, without the need of full CFD simulations [5]. By offering rapid, data-driven predictions, deep learning techniques can significantly accelerate the airfoil design process while maintaining high accuracy. Several studies have explored the application of deep learning in airfoil design and performance prediction [6].

[†]Joint first authors
[*]Corresponding author: vks.ade@gov.in

Sekar *et al.* [7] introduced an inverse design framework utilizing Deep Convolutional Neural Networks to predict airfoil geometries based on target aerodynamic performance parameters. Their model demonstrated considerable improvements in both accuracy and computational efficiency. Duru *et al.* [8] applied a deep learning approach to predict transonic flow fields around airfoils, showcasing the ability of neural networks to capture complex aerodynamic phenomena typically resolved through CFD methods [9]. Kim and Yoon [10] proposed a deep CNN model aimed at enhancing airfoil performance through geometric modifications. Their work emphasizes the utility of neural networks in optimizing airfoil shapes for im- proved aerodynamic efficiency. Similarly, Li *et al.* [11] employed a deep learning framework for low *Re* airfoil design, utilizing tailored airfoil modes to optimize performance. Their approach achieved significant accuracy in predicting aerodynamic characteristics and showed promise for practical application in airfoil design optimization. Sun *et al.* [12] expanded the scope of in- verse design by using Artificial Neural Network (ANN) to predict both airfoil and wing configurations, providing a comprehensive design framework that generalizes well across different aerodynamic surfaces. In another work, Yilmaz and German [13] employed CNNs to predict airfoil performance based on geometric input, while Zhang *et al.* [14] developed a CNN approach to predict lift coefficients of airfoils [15], highlighting the potential for deep learning models possibly replace traditional CFD simulations in of the applications. Li *et al.* [16] further explored a data-driven approach in transonic aerodynamic shape optimization, ensuring low-speed performance through deep learning constraints. Du *et al.* [17] advanced airfoil design by incorporating neural network-based parameterization and surrogate modeling, enabling rapid design optimization. Their work significantly reduced computational costs while maintaining accurate predictions. Chen *et al.* [18] extended this application by using CNNs to predict multi-aerodynamic coefficients for airfoils, enhancing the scope of airfoil design tasks that can be automated through deep learning. Additionally, Hui *et al.* [19] developed a fast pressure distribution prediction model using deep learning, emphasizing the speed at which deep learning techniques can produce reliable aerodynamic data for airfoil designs [20]. In a recent study, the effectiveness of CNN and DNN models are compared for the inverse design of airfoil shapes based on pressure distributions [21]. The CNN model used images of Coefficient of Pressure ($C_p$) distributions, while the DNN model utilized numerical $C_p$ data, both trained on data generated from XFoil [22]. Hyper parameters such as learning rate, batch size, and epochs were optimized to minimize Mean Square Error (MSE) and improve prediction accuracy, with the DNN model proving approximately seven times faster and more accurate than the CNN model [21]. The airfoils are characterized using 160 panels, clustered at the leading and trailing edges. Datasets are generated with the *Re* varied from 60,000 to 120,000 and *AoA* ranging from 5° to 20°. For the analysis, *Re* of 100,000 is selected, with an *AoA* of 3° for the CNN and 5° for the DNN, and results from the CNN at 5° we included in the validation section for direct comparison with the DNN model [21]. In the present proposed model, the work of [21] is further extended by taking into consideration *AoA*, *Re*, and several other key improvements in airfoil design using deep learning techniques. The present work consists of prediction of $C_p$ distributions from airfoil shapes as well as inverse of it, establishes a bidirectional design pipeline. Additionally, extensive datasets are generated by increasing *Re* range from 10,000 to 9,000,000 and varying *AoA* from 0° to 15°, compared to the limited dataset [21]. The performance of model is evaluated against multiple cases incorporating variations in *AoA* and *Re* under realistic aerodynamic conditions, demonstrating the robustness of the present approach. In the last, a flying wing aircraft model of wing span > 10 m is considered for the prediction of pressure distribution along the chord wise and the CNN and DNN models show promising results. This research underscores the potential of deep learning models accelerating aerodynamic optimization and advancing the design of high-performance airfoil design and development using computationally efficient frameworks. Details of datasets are presented in section 2. Architecture of Neural network models are presented in section 3. Results and discussions are presented in section 4 followed by concluding remarks in section 5.

## 2. DATASET

The datasets used in the present study are generated using Xfoil [22]. Xfoil employs a high-order panel method combined with a fully coupled viscous/inviscid interaction technique to assess various boundary layer parameters. Xfoil takes airfoil coordinates as input along with specific boundary conditions like *AoA*, *Re*, Mach Number (*M*) and produces pressure distributions on the top and bottom surfaces of airfoil. In the present work, *AoA* is varied from 0 to 15 degree with a step of 1 degree and the *Re* varied between 10,000 and 90,00,000 for a given *M* of 0.5. A total number of 1609 open-source airfoils are considered in the present study, as shown in Figure 1 with varying *AoA*. All the airfoil coordinates are stored in .DAT format, which was later employed to train both CNN and DNN models. $C_p$ distributions of all the airfoil are generated with varying $Re$ and $AoA$. The number of panels was set to 125 for all airfoils, with increased density near the leading edge to accommodate greater curvature.

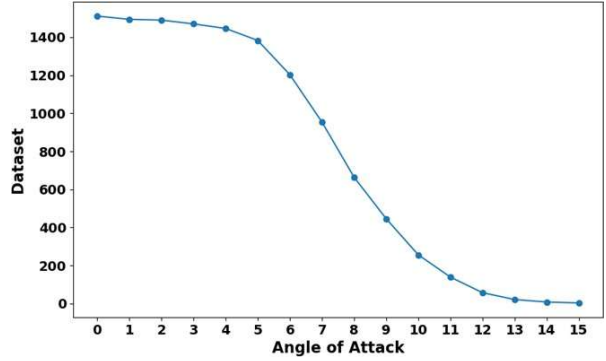

**FIGURE 1: NUMBER OF DATASETS WITH VARYING $AoA$ FOR $Re = 2 \times 10^6$**

## 3. NEURAL MODEL AND MODEL ARCHITECTURE

In the present paper, two approaches, i.e., CNN and DNN is used. In conventional machine learning, pressure distribution is predicted for a given airfoil. In the present study either (a) pressure distributions (output as coordinates) are predicted for a given airfoil shape (input given as coordinate, in the case of DNN and image, in the case of CNN), as shown in Fig 2. or (b) airfoil shapes (output as coordinates) are predicted for a given pressure distribution (input given as coordinate, in the case of DNN and image, in the case of CNN). The CNN and DNN models are created in Python using Jupyter Notebook [23]. These trained models are subsequently used to predict either airfoil shapes or $C_p$ distributions.

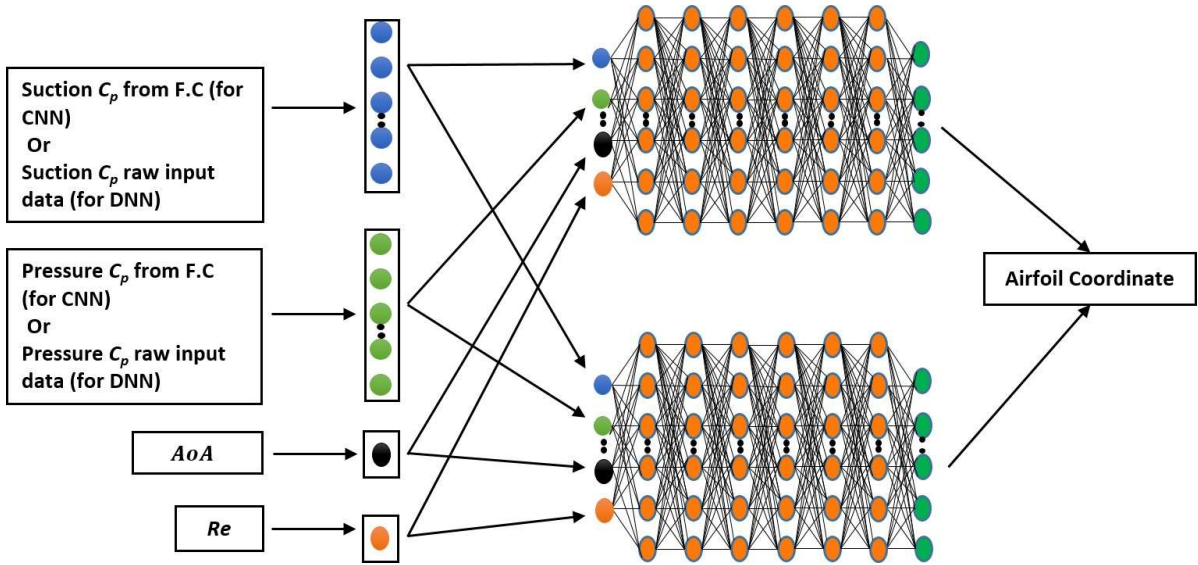

**FIGURE 2: CNN/DNN MODEL ARCHITECTURE**

### 3.1. Convolution Neural Network

CNN model is developed using the image data as input [24]. CNN model consists of convolutional layers, pooling layer, and a Fully Connected Layer (F.C) (refer Table 1 for further details) [25]. The image fed to the CNN model passes through these layers, which help the model to enhance the important characteristics of an image. OpenCV library of python is used for importing and read the images [26]. The data is divided into training data and testing data with an 80:20 split. The training data is fed to the CNN Model for learning. The details of convolution layers and MaxPooling including the parameters like figure size, size of kernel matrix, number of stride and type of activation function is shown in Table 1. The input $C_p$ figure size of 200×200 is considered. The designed Convolutional Neural Network architecture has three 2D convolutional layers each of them followed by a maximum 2D pooling layer. The matrix generated is then flattened and is given as input to Neural Network with three hidden layers and one output layer. A linear activation function is used for the output layer since it's a regression type data. For the rest of the layers, Rectified Linear Unit (ReLU) is used as the activation function to capture non-linearity in the data. The output matrix size is 250×1 with the first 125 elements corresponding to x-coordinates, and the rest of the 125 elements correspond to y-coordinates of a given airfoil.

**TABLE 1: PARAMETERS FOR CONVOLUTIONAL LAYERS**

| Type | Maps | Figure Size | Kernel | Stride | Activation |
|---|---|---|---|---|---|
| Input | 1 | 200×200 | - | - | - |
| Conv2D | 32 | 198×198 | (3×3) | 1 | ReLU |
| MaxPooling | 1 | 99×99 | (2×2) | 1 | - |
| Conv2D | 64 | 97×97 | (3×3) | 1 | ReLU |
| MaxPooling | 1 | 48×48 | (2×2) | 1 | - |
| Conv2D | 128 | 46×46 | (3×3) | 1 | ReLU |
| MaxPooling | 1 | 23×23 | (2×2) | 1 | - |
| F.C | - | 67312 | - | - | - |

**TABLE 2: PARAMETERS FOR DNN MODEL**

| Parameter | Value |
|---|---|
| Input Size | 252 (125, 125, 1, 1) |
| Output Size | 250 (125, 125) |
| Architecture | [125, 250, 236, 375, 250] |
| Activation | ReLU |
| Batch Size | 32 |
| Epochs | 500 |
| Learning Rate | $1\times10^{-4}$ |
| Loss Function | MSE |

### 3.2. Deep Neural Network

A DNN model works on numerical data in the form of a vector or a matrix. These matrices are processed through different set of dense layers and different weights assigned to each layer at initial. DNN model is trained through iterative methods. Input data ($C_p$) for DNN is also divided into two sections, i.e., pressure and suction. To minimize the error, weights are updated in each iteration which results in decreasing the loss. Table 2 illustrates the specifications of the DNN model developed for predicting the airfoil coordinates. All the layers utilized ReLU activation function which helps in linear and nonlinear relationships and prevent vanishing gradient issues. The output layer consists of 250 neurons apply a linear activation function to predict the airfoil (or $C_p$). Figure 2 shows the CNN architecture, where the outcome of fully connected layers for $C_p$ suction and pressure along with *AoA* and *Re* is passed through Neural Network to predict airfoil coordinates as well as DNN architecture, where $C_p$ suction and pressure raw data along with *AoA* and *Re* is used as input for predicting airfoil coordinates.

**TABLE 3: INPUT (IMAGE FOR CNN AND COORDINATES FOR DNN) AND OUTPUT (COORDINATES FOR CNN AND DNN) DESCRIPTION FOR EACH TEST CASE**

| Sr. No. | Re | AoA | Input | Output |
|---|---|---|---|---|
| 1 | Constant | Constant | $C_p$ and AoA | Airfoil |
| 2(a) | Constant | Variable | $C_p$ and AoA | Airfoil |
| 2(b) | Constant | Variable | $C_p$ and AoA | Airfoil |
| 3 | Variable | Variable | $C_p$, AoA, and Re | Airfoil |
| 4(a) | Constant | Variable | Airfoil and AoA | $C_p$ |
| 4(b) | Constant | Variable | Airfoil and AoA | $C_p$ |
| 5 | Variable | Variable | Airfoil, AoA, and Re | $C_p$ |

## 4. TEST CASES

In the present study, multiple test cases are considered. For each case, both CNN and DNN models are studied. The details of test cases are also presented in Table 3.

**Case 1:** In Case 1, airfoils are analyzed at a zero AoA for the given constant Re of $2\times10^6$ and M = 0.5. Data contains 1,609 airfoils, with $C_p$ for each airfoil as input. This data is used to create CNN and DNN models to get airfoil coordinates at output.

**Case 2(a):** In this case, AoA is varied from 0° to 15° (keeping the Re constant of $2\times10^6$ and for given M = 0.5) with an increment of 1°. Variation in AoA is achieved by altering free stream airflow direction. The dataset includes 10,970 cases, with pressure distributions recorded at each AoA increment. Both the CNN and DNN models are developed to predict airfoil coordinates as output using $C_p$ suction, $C_p$ pressure, and AoA as inputs.

**Case 2(b):** In this case, AoA is varied from 0° to 15° (keeping Re constant for $2\times10^6$ and for given M = 0.5). Variation in AoA is achieved by physically rotating the airfoil while maintaining a constant airflow direction. Each airfoil was rotated to the desired AoA, and subsequently, XFoil simulations are carried out to generate $C_p$ distributions, resulting in a dataset of 10,970 airfoil cases. Both the CNN and DNN models are developed to predict airfoil coordinates as output using $C_p$ suction, $C_p$ pressure, and AoA as inputs.

**Case 3:** In this case, both the CNN and DNN models are developed to predict airfoil coordinates as output using $C_p$ suction, $C_p$ pressure, AoA, and Re as inputs. A dataset of 1,06,366 airfoil cases is considered ensuring a more diverse learning framework for the model.

**Case 4(a):** In this case, the same datasets of Case-2(a) is used (i.e., for a constant Re of $2\times106$ with varying AoA from 0° to 15°). However, the inputs and outputs for CNN and DNN models are interchanged. For each model, input is considered as an airfoil and AoA for the prediction of $C_p$ suction and $C_p$ pressure coordinates as output.

**Case 4(b):** In this case, the same datasets of Case-2(b) is used (i.e., for a constant Re of $2\times10^6$ with varying AoA from 0° to 15°). However, the inputs and outputs for CNN and DNN models are interchanged. For each model, input is considered as an airfoil and AoA for the prediction of $C_p$ suction and $C_p$ pressure coordinates as output.

**Case 5:** In this case, the same datasets of Case-3 are used (i.e., varying Re and AoA). However, the inputs and outputs for CNN and DNN models are interchanged. For each model, input is considered as an airfoil, AoA, and Re for the prediction of $C_p$ suction and $C_p$ pressure coordinates as output.

## 5. RESULTS AND DISCUSSIONS
### 5.1. CNN

Firstly, airfoil coordinate prediction using the CNN model for Case – 2(a) is shown in Figure 3. True and predicted airfoil coordinates of a typical airfoil, i.e., FX 72-LS-160 airfoil at AoA = 9° are shown in Figure 3(a). The subsequent percentage error of FX 72-LS-160 airfoil at AoA = 9° is shown in Figure 3(b). The result shows an error of up to 12% (locally) for the FX 72-LS- 160 airfoil at $AoA = 9°$. True and predicted airfoil coordinates of another airfoil, i.e., EPPLER 333 airfoil at $AoA = 6°$ are also shown in Figure 3(c) for Case 2(b). The percentage error of EPPLER 333 airfoil at $AoA = 6°$ using CNN is shown in Figure 3(d). The result shows the localized error of 8% at the pressure side. However, an overall average percentage errors of 7% and 9% are obtained for Case – 2(a) and Case – 2(b), respectively. Training losses of 5.5E-6 and 8E-6 and testing losses of 2.83E-4 and 2.96E-4 are obtained for Cases 2(a) and 2(b), respectively. Airfoil coordinates prediction using the CNN model for Case – 3 are shown in Figure 4. In this case, input to the CNN model consists of $Re$, $AoA$, $C_p$ suction, and $C_p$ pressure distributions. The model is trained across 27 different $Re$ values to capture the variations in aerodynamic behavior and enhance its generalization capability. True and predicted airfoil coordinates of an airfoil, i.e., EPPLER-598 airfoil at $Re = 5E+06$ and $AoA = 8°$ are shown in Figure 4(a). The subsequent percentage error of EPPLER 598 airfoil at $Re = 5E+06$ and $AoA = 8°$ is shown in Figure 4(b). Result shows an 8% error (locally) for the EPPLER-598 airfoil at $Re = 5E+06$ and $AoA = 8°$. However, an overall average percentage error of 9% is obtained for Case – 3 respectively. Training and testing losses of 6E-4 and 4.3E-3 are obtained for Case – 3. An example of Case-4(a) is shown in Figure 5(a), where the input to the model is airfoil coordinates along with $AoA$ and the output is $C_p$ distribution. The predicted $C_p$ values are compared with the true values, as shown in Figure 5(a) for W1011 airfoil for a given $AoA = 2°$. Subsequently, error plots are generated for both the suction and pressure surfaces, as shown in Figure 5(b). A maximum error of 8% is observed. Airfoil GOE 174 is considered for demonstrating Case – 4(b) for a given $AoA = 4°$. The predicted $C_p$ values are compared with the true values, as shown in Figure 5(c) for GOE 174 airfoil for a given $AoA = 4°$. Subsequently, error plots are generated for both the suction and pressure surfaces, as shown in Figure 5(d). A maximum error of 10% is observed. However, overall average percentage errors of 8.5% and 10% are obtained for Case – 4(a) and Case – 4(b), respectively. Training losses of 4.3E-5 and 5.3E-5 and testing loss of 3.1E-4 and 4.2E-4 are obtained for Case 4(a) and 4(b), respectively. Finally, $C_p$ coordinates are predicted using the CNN model for Case – 5. In this case, input to the CNN model consists of airfoil, $Re$, and $AoA$. The model is trained across 27 different $Re$ values to capture the variations in aerodynamic behavior and enhance its generalization capability. True and predicted $C_p$ distribution of NACA 2415 airfoil at $Re = 2E+04$ and $AoA = 9°$ is shown in Figure 6(a). The subsequent percentage error of NACA 2415 airfoil at $Re = 2E+04$ and $AoA = 9°$ is shown in Figure 6(b). Result shows 12% error (locally) for the NACA 2415 airfoil at $Re = 2E+04$ and $AoA = 9°$. However, an overall average percentage error of 9.5% is obtained for Case – 5 respectively. Training and testing losses of 6E-4 and 4.3E-3 are obtained for the Case – 5.

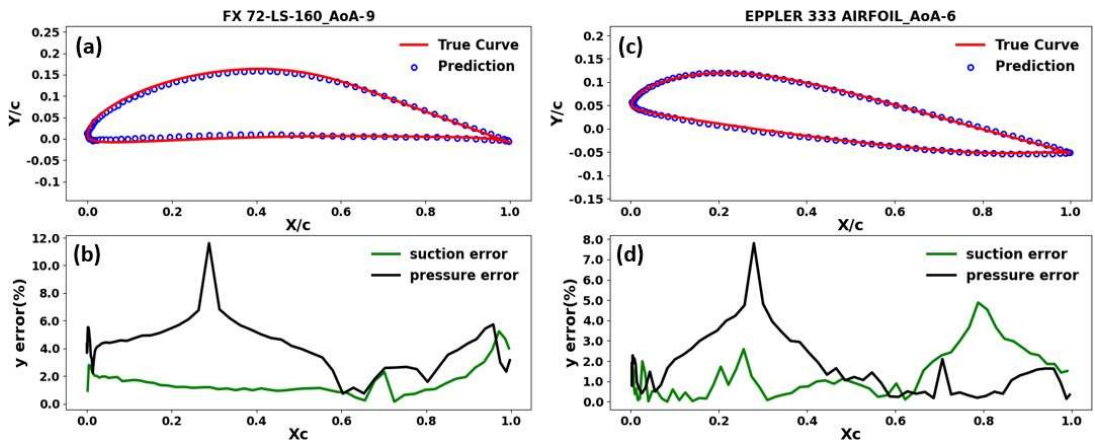

**FIGURE 3: TRUE AND PREDICTED AIRFOIL COORDINATES OF (A) FX 72-LS-160 AIRFOIL AT *AoA* = 9° (B) PERCENTAGE ERROR OF FX 72-LS-160 AIRFOIL AT *AoA* = 9° FOR CASE 2(A) USING CNN, (C) EPPLER 333 AIRFOIL AT *AoA* = 6°, AND (D) PERCENTAGE ERROR OF EPPLER 333 AIRFOIL AT *AoA* = 6° FOR CASE 2(B) USING CNN MODEL**

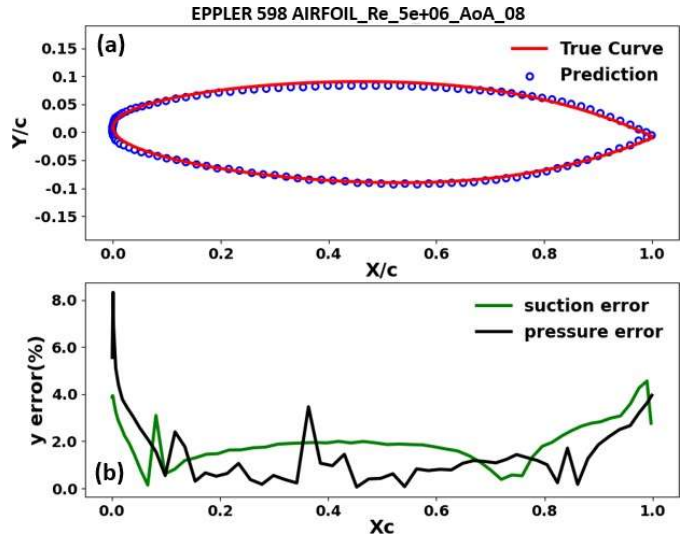

**FIGURE 4: TRUE AND PREDICTED AIRFOIL COORDINATES OF (A) EPPLER 598 AIRFOIL AT *Re* = 5E+06 AND *AoA* = 8° AND (B) PERCENTAGE ERROR OF EPPLER 598 AIRFOIL AT *Re* = 5E+06 AND *AoA* = 8° FOR CASE 3 USING CNN MODEL**

### 5.2. DNN

Airfoil coordinate prediction using DNN model for Case 2(a) is shown in Figure 7. True and predicted airfoil coordinates of a typical airfoil i.e., S1091 airfoil at *AoA* = 0° are shown in Figure 7(a). The Subsequent percentage error of S1091 airfoil at *AoA* = 0° is shown in Figure 7(b). Results show the error up to 7% error (locally) for S1091 airfoil at *AoA* = 0°. True and predicted airfoil coordinates of another airfoil, i.e., GOE 284 airfoil at *AoA* = 11° are also shown in Figure 7(c) for Case 2(b). The percentage error of GOE 284 airfoil at *AoA* = 11° using DNN is shown in Figure 7(d). The result shows a localized error of 7% at pressure side. However, an overall average percentage errors of 5.5% and 6% are obtained for Case – 2(a) and Case – 2(b), respectively. Training losses of 9.0E-6 and 7.0E-4 and testing losses of 3.0E-5 and 3E-3 are obtained for Case 2(a) and 2(b), respectively. Airfoil coordinate prediction using DNN model for Case 3 is shown in Figure 8. In this case, input to the DNN model consists of *Re*, *AoA*, $C_p$ suction, and $C_p$ pressure distributions. The total validation loss computed as combined loss for both suction and pressure surfaces, converged at 2.1E-5. The model is trained for 500 epochs. True and predicted airfoil coordinates of an airfoil, i.e., CURTISS CR-1 airfoil at *Re* = 3E+04 and *AoA* = 10° are shown in Figure 8(a).

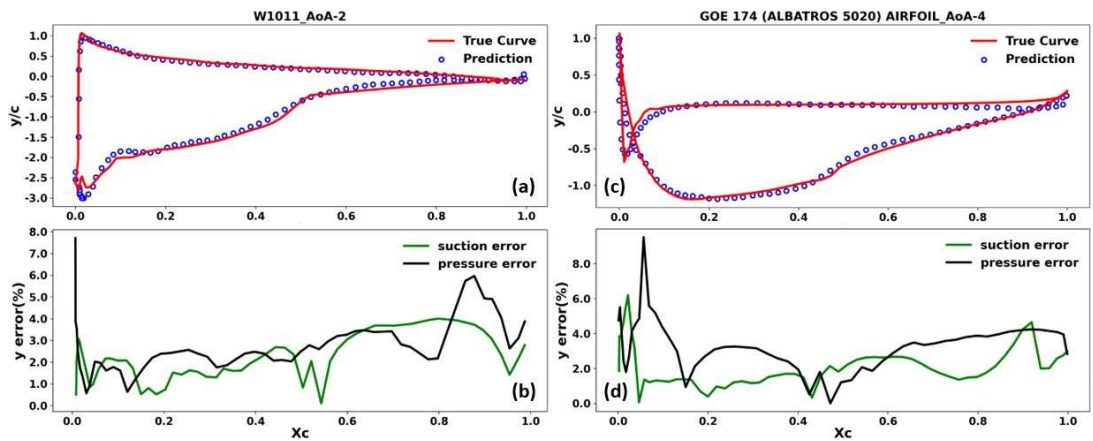

**FIGURE 5: TRUE AND PREDICTED AIRFOIL COORDINATES OF (A) W1011 AIRFOIL AT *AoA* = 2° (B) PERCENTAGE ERROR OF W1011 AIRFOIL AT *AoA* = 2° FOR CASE 4(A) USING CNN, (C) GOE 174 AIRFOIL AT *AoA* = 4°, AND (D) PERCENTAGE ERROR OF GOE 174 AIRFOIL AT *AoA* = 4° FOR CASE 4(B) USING CNN MODEL**

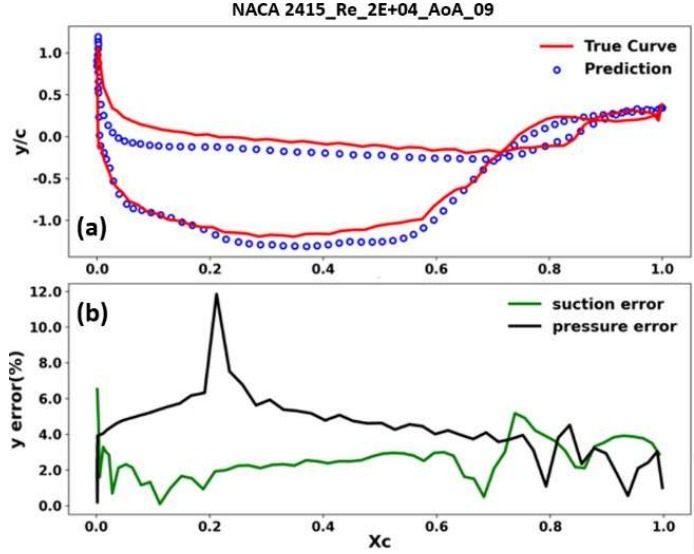

**FIGURE 6: TRUE AND PREDICTED AIRFOIL COORDINATES OF (A) NACA 2415 AIRFOIL AT *Re* = 2E+04 AND *AoA* = 9° AND (B) PERCENTAGE ERROR OF NACA 2415 AIRFOIL AT *Re* = 2E+04 AND *AoA* = 9° FOR CASE 5 USING CNN MODEL**

The subsequent percentage error of CURTISS CR-1 airfoil at $Re$ = 3E+04 and $AoA$ = 10° is shown in Figure 8(b). Result shows error up to 5% (locally) for the CURTISS CR-1 airfoil at $Re$ = 3E+04 and $AoA$ = 10°. However, an overall average percentage error of 7.5% is obtained for Case – 3. Training and testing losses of 2E-5 and 2.23E-4 are obtained for Case - 3. An example of Case-4(a) is shown in Figure 9(a), where the input to the model is airfoil coordinates along with $AoA$ and the output is $C_p$ distribution. The predicted $C_p$ values are compared with the true values, as shown in Figure 9(a) for SIKORSKY SC1095 airfoil for a given $AoA$ = 6°. Subsequently, error plots are generated for both the suction and pressure surfaces, as shown in Figure 9(b). A maximum error of 8% is observed. Airfoil MH112 16.2% is considered for demonstrating Case 4(b) for a given $AoA$ = 11°. The predicated $C_p$ values are compared with the true values, as shown in Figure 5(c) for MH112 16.2% for a given $AoA$ = 11°. Subsequently, error plots are generated for both the suction and pressure surfaces, as shown in Figure 5(d). A maximum error of 8% is observed. However, an overall average percentage errors of 8% and 8.5% is obtained for Case – 4(a) and Case – 4(b), respectively. Training losses of 1.3E-5 and 2.3E-5 are obtained for Cases 4(a) and 4(b), respectively. Testing losses of 6.3E-4 and 7.2E-4 for Cases 4(a) and 4(b), respectively. Finally, $C_p$ coordinates are predicted using the DNN model for Case – 5. In this case, input to the DNN model consists of airfoil, $Re$, and $AoA$. The model is trained across 27 different $Re$ values to capture the variations in aerodynamic behavior and enhance its generalization capability. True and predicted $C_p$ distributions of BOEING BACXXX airfoil at $Re$ = 3E+05 and $AoA$ = 4° are shown in Figure 10(a). The subsequent percentage error of BOEING BACXXX airfoil at $Re$ = 3E+05 and $AoA$ = 4° is shown in Figure 10(b). Result shows the error up to 12% (locally) for the BOEING BACXXX airfoil at $Re$ = 3E+05 and $AoA$ = 4°. However, an overall average percentage error of 7.5% is obtained for Case – 5. Training and testing losses of 5.2E-4 and 3.5E-3 are obtained for Case-5.

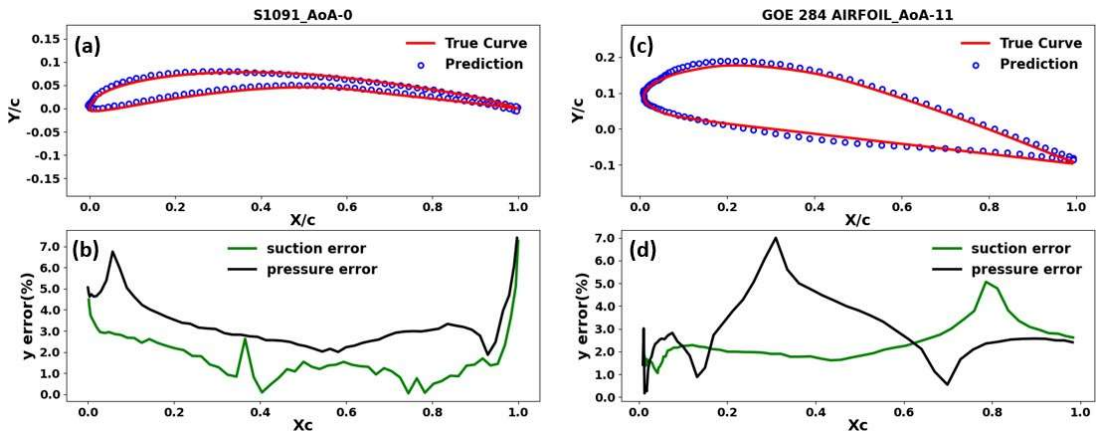

FIGURE 7: TRUE AND PREDICTED AIRFOIL COORDINATES OF (A) S1091 AIRFOIL AT $AoA$ = 0° (B) PERCENTAGE ERROR S1091 AIRFOIL AT $AoA$ = 0° FOR CASE 2(A) USING DNN, (C) GOE 284 AIRFOIL AT $AoA$ = 11°, AND (D) PERCENTAGE ERROR OF GOE 284 AIRFOIL AT $AoA$ = 11° FOR CASE 2(B) USING DNN MODEL

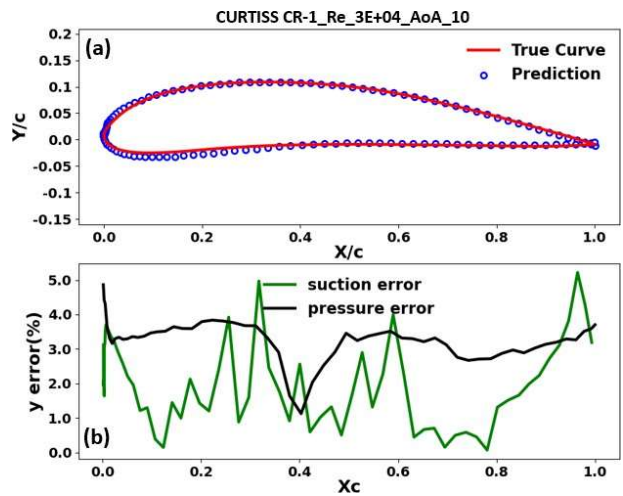

FIGURE 8: TRUE AND PREDICTED AIRFOIL COORDINATES OF (A) CURTISS CR-1 AIRFOIL AT $Re$ = 3E+04 AND $AoA$ = 10° AND (B) PERCENTAGE ERROR OF CURTISS CR-1 AIR- FOIL AT $Re$ = 3E+04 AND $AoA$ = 10° FOR CASE 3 USING DNN MODEL

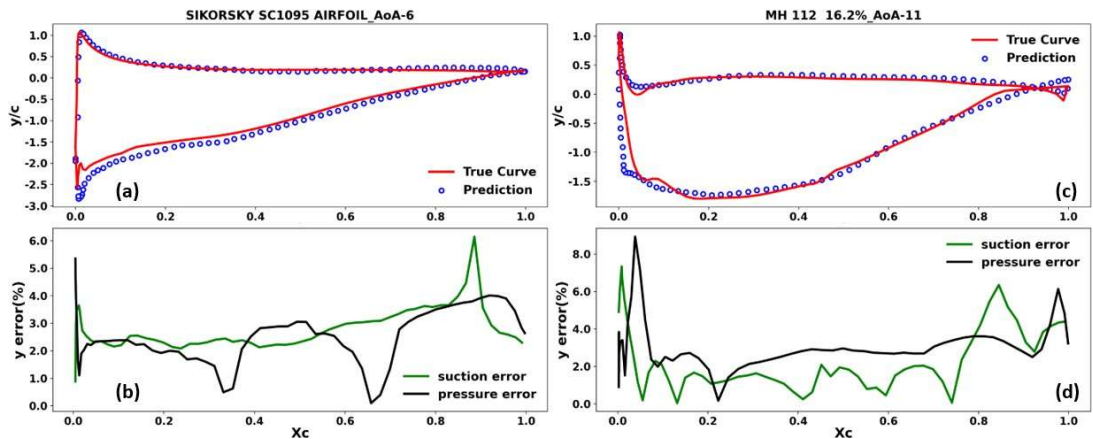

FIGURE 9: TRUE AND PREDICTED AIRFOIL COORDINATES OF (A) SIKORSKY SC1095 AIRFOIL AT $AoA$ = 6° (B) PERCENTAGE ERROR OF SIKORSKY SC1095 AIRFOIL AT $AoA$ = 6° FOR CASE 4(A) USING DNN, (C) MH112 16.2% AIRFOIL AT $AoA$ = 11°, AND (D) PERCENTAGE ERROR OF MH112 16.2% AIRFOIL AT $AoA$ = 11° FOR CASE 4(B) USING DNN MODEL

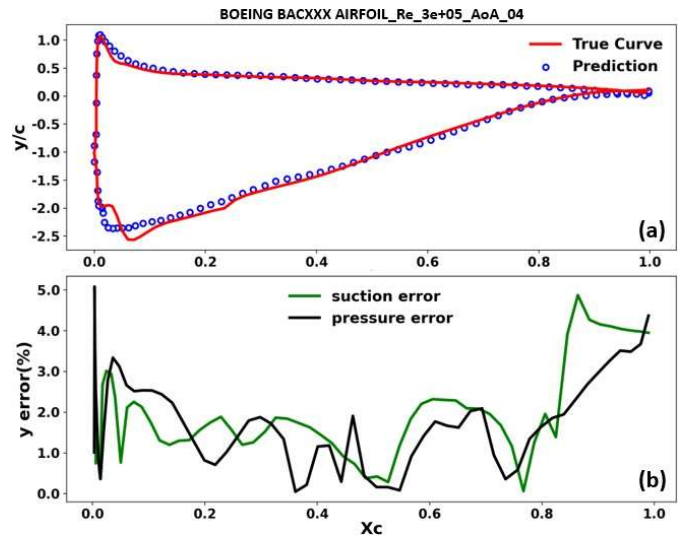

FIGURE 10: TRUE AND PREDICTED AIRFOIL COORDINATES OF (A) BOEING BACXXX AIRFOIL AT *Re* = 3E+05 AND *AoA* = 4° AND (B) PERCENTAGE ERROR OF BOEING BACXXX AIR- FOIL AT *Re* = 3E+05 AND *AoA* = 4° FOR CASE 5 USING DNN MODEL

## 6. CONCLUSIONS

The current study has demonstrated the effectiveness of deep learning models, specifically Convolutional Neural Networks (CNNs) and Deep Neural Networks (DNNs), in predicting airfoil geometries from target pressure distributions and vice versa. The integration of varying *Re* and *AoA* into the training process significantly enhanced the generalization capability of the models. The CNN-based approach effectively captured spatial dependencies within pressure distributions, while the DNN model provided a computationally efficient alternative with comparable accuracy. The results indicate that deep learning methods can serve as viable alternatives to traditional CFD simulations, enabling faster aerodynamic optimization. Potential modifications to further enhance the model's performance include incorporating Mach Number and angle of side slip as additional input variables. Additionally, fine-tuning hyperparameters such as learning rate, batch size, and the number of layers could further help minimize loss and improve prediction accuracy. These enhancements would contribute to a more robust and efficient framework for accelerated airfoil design, making deep learning an increasingly valuable tool in aerodynamic analysis and optimization.

## ACKNOWLEDGMENTS

Authors would like to express our profound gratitude to Shri Y Dilip, Director of Aeronautical Development Establishment, DRDO (ADE), Mr. Manjunath S M, Technology Director (TD) and Mr. Diptiman Biswas, Group Director (GD) of ADE, DRDO.